\documentclass[conference]{IEEEtran}
\IEEEoverridecommandlockouts
\usepackage{cite}
\usepackage{amsmath,amssymb,amsfonts,amsthm}
\usepackage{algorithmic}
\usepackage{graphicx}
\usepackage{textcomp}
\usepackage{xcolor}
\usepackage{bm}
\usepackage{comment}

\theoremstyle{definition}
\newtheorem{theo}{Theorem}
\newtheorem{defi}{Definition}
\newtheorem{lem}{Lemma}

\newtheorem{coro}{Corollary}

\newtheorem{proofl}{Proof of Lemma}
\newtheorem{prooft}{Proof of Theorem}
\newtheorem{proofc}{Proof of Corollary}
\newtheorem{ex}{Example}

\def\BibTeX{{\rm B\kern-.05em{\sc i\kern-.025em b}\kern-.08em
    T\kern-.1667em\lower.7ex\hbox{E}\kern-.125emX}}
\begin{document}

\title{Theoretical Analysis of the Advantage of\\ Deepening Neural Networks
}

\author{
\IEEEauthorblockN{Yasushi Esaki}
\IEEEauthorblockA{\textit{Faculty of Science and Engineering} \\ 
\textit{Waseda University}\\
Tokyo, Japan \\
esaki@fuji.waseda.jp}
\and

\IEEEauthorblockN{Yuta Nakahara}
\IEEEauthorblockA{\textit{Center for Data Science} \\
\textit{Waseda University}\\
Tokyo, Japan \\
yuta.nakahara@aoni.waseda.jp}
\and

\IEEEauthorblockN{Toshiyasu Matsushima}
\IEEEauthorblockA{\textit{Faculty of Science and Engineering} \\
\textit{Waseda University}\\
Tokyo, Japan \\
toshimat@waseda.jp}
}

\maketitle

\begin{abstract}
We propose two new criteria to understand the advantage of deepening neural networks. It is important to know the expressivity of functions computable by deep neural networks in order to understand the advantage of deepening neural networks. Unless deep neural networks have enough expressivity, they cannot have good performance even though learning is successful. In this situation, the proposed criteria contribute to understanding the advantage of deepening neural networks since they can evaluate the expressivity independently from the efficiency of learning. The first criterion shows the approximation accuracy of deep neural networks to the target function. This criterion has the background that the goal of deep learning is approximating the target function by deep neural networks. The second criterion shows the property of linear regions of functions computable by deep neural networks. This criterion has the background that deep neural networks whose activation functions are piecewise linear are also piecewise linear. Furthermore, by the two criteria, we show that to increase layers is more effective than to increase units at each layer on improving the expressivity of deep neural networks.
\end{abstract}


\section{Introduction}
\label{secintro}
In recent researches on machine learning, deep neural networks show very good performance on prediction. For example, some papers consider that to increase layers in neural networks contributes to the accuracy of image recognition e.g., \cite{Simonyan, Veit}. There are also papers which consider that to increase layers contributes to the accuracy of speech recognition e.g., \cite{Hochreiter, Hochreiter97}. Neural networks are getting deeper and deeper with each passing year. 

In this situation, there are many researchers that are interested in the relation between the depth of neural networks and good performance. In order to show the advantage of deepening neural networks, the followings should be distinguished.
\begin{enumerate}
\item the expressivity of deep neural networks as functions which express the relation between explanatory variables and objective variables.
\item the efficiency of learning by the gradient method.
\end{enumerate}
The prediction accuracy after learning is dependent on both the expressivity and the efficiency of learning. Unless deep neural networks have enough expressivity, they cannot have good performance even though learning is successful.

Therefore, there have been some researches that evaluate deep neural networks with distinguishing between the expressivity and the efficiency of learning e.g., \cite{Yarotsky18, Telgarsky16}. Some of the researches assume that data is scattered around the target function and investigate the approximation accuracy when approximating the target function by deep neural networks. They discuss the distance between deep neural networks and the target function. Then they regard that the shorter that distance is, the more expressivity the deep neural networks have. For example, some studies consider how many layers and units are necessary for the distance between deep neural networks and the target function to be less than $\epsilon$, for any $\epsilon>0$ and any target function e.g., \cite{Cybenko, Hornik}. Besides, other studies consider how the lower limit of the distance between deep neural networks and the target function is described for the depth and the number of units e.g., \cite{Mhaskar, Yarotsky}.

On the other hand, there are other kinds of researches that evaluate deep neural networks with distinguishing between the expressivity and the efficiency of learning. Those researches investigate the property of linear regions of functions by computable by deep neural networks. Since deep neural networks whose activation functions are piecewise linear are also piecewise linear, investigating the property of linear regions is a good way to know the shape of the lines drawn by those deep neural networks. For example, some studies count the number of linear regions of deep neural networks \cite{Montufar, Pascanu, Serra}. They regard that the larger the number of linear regions of deep neural networks is, the more expressivity the deep neural networks have. Besides, other studies calculate the volume of the boundary between two linear regions of deep neural networks \cite{Hanin}. They regard that the larger the volume of the boundary between two linear regions of deep neural networks is, the more expressivity the deep neural networks have. 

In this paper, we propose two criteria to evaluate the expressivity of deep neural networks independently from the efficiency of learning, similar to the above studies. The first criterion evaluates the approximation accuracy of deep neural networks to the target function. It shows the ratio of the values of the parameters which make the distance between deep neural networks and the target function short. We regard that the larger the ratio is, the more expressivity deep neural networks have. The novelty of this criterion is that we can figure out the volume of parameter values which give functions close to the target function. The proposed criterion shows not only the existence of the appropriate parameter values but also how many the appropriate parameter values exist.

The second criterion evaluates the property of linear regions of functions computable by deep neural networks whose activation functions are piecewise linear. It shows the size of the maximal linear region of deep neural networks. We regard that the smaller the size of the maximal linear region is, the more expressivity the deep neural networks have. The novelty of this criterion is figuring out how fine linear regions the area which is the least flexible has. The proposed criterion shows whether linear regions are scattered finely and uniformly or not.

\begin{figure}[t]
\begin{center}
\centerline{\includegraphics[height=3.5cm,width=\linewidth]{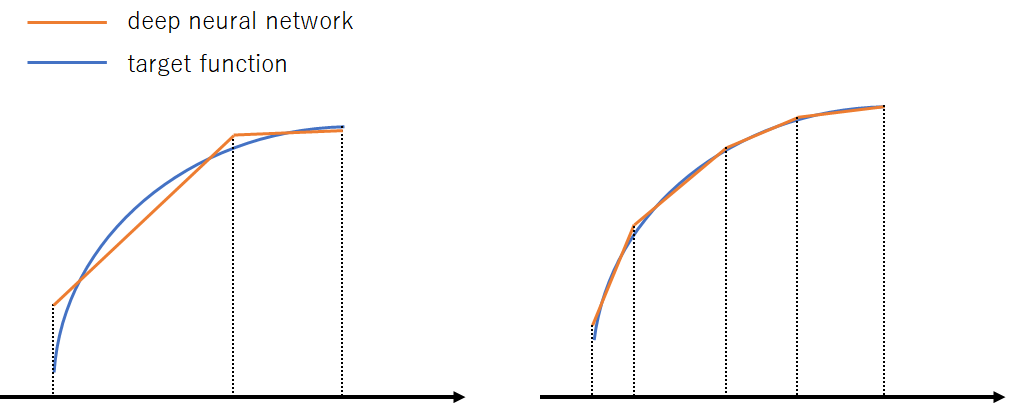}}
\caption{The relation between the size of the maximal linear regions of deep neural networks and the approximation to the target function by deep neural networks. If the size of the maximal linear region of deep neural networks is large, a gap inevitably occurs between deep neural networks and the target function, as the left figure. On the other hand, if the size of the maximal linear region of deep neural networks is small, there is a possibility that the target function is approximated by deep neural networks, as the right figure.}
\label{figlinearappro}
\end{center}
\end{figure}

The above two criteria have a relation. The small size of the maximal linear region of deep neural networks is a requirement for the good approximation to the target function if the target function is smooth. This requirement is understood by referring to Fig. \ref{figlinearappro}. Unless the size of the maximal linear region of deep neural networks is small, there is a gap between the deep neural networks and the target function no matter how close the deep neural networks and the target function.

Moreover, by the two criteria, we show that to increase layers is more effective than to increase units at each layer on improving the expressivity of neural networks. We adopt theoretical proofs and computer simulations to show it.

In summary, our contributions are as follows.
\begin{enumerate}
\item We propose new criteria to evaluate the expressivity of deep neural networks independently from the efficiency of learning.
\item We show that to increase layers is more effective than to increase units at each layer on improving the expressivity of neural networks, by the new criteria.
\end{enumerate}
\section{Definitions and Theoretical Facts}
\subsection{Deep Neural Networks}
\label{secdeep}
We define a neural network as a composite function $\bm{F}_{\bm{\theta}} : \mathbb{R}^{n_0}\to\mathbb{R}^{n_L}$ parametrized by $\bm{\theta}$, which alternately computes linear functions $\bm{f}^{(l)}: \mathbb{R}^{n_{l-1}}\to\mathbb{R}^{n_l}\ (l=1,\cdots,L)$ and non-linear activation functions $\bm{g}^{(l)} : \mathbb{R}^{n_l}\to\mathbb{R}^{n_l}\ (l=1,\cdots,L-1)$ such as
\begin{align}
\bm{F}_{\bm{\theta}}=\bm{f}^{(L)}\circ \bm{g}^{(L-1)}\circ \bm{f}^{(L-1)}\circ\cdots\circ \bm{g}^{(1)}\circ \bm{f}^{(1)}. \nonumber
\end{align}

The linear function $\bm{f}^{(l)}$ is given by 
\begin{align}
f^{(l)}_j(\bm{x})=\sum^{n_{l-1}}_{i=1}w^{(l)}_{ji}x_i+b^{(l)}_j\  (j=1,\cdots,n_l), \nonumber 
\end{align}
where $x_i$ is the $i$-th coordinate of $\bm{x}$ and $f^{(l)}_j$ is the $j$-th coordinate of $\bm{f}^{(l)}$. The parameter $\bm{\theta}$ is a vector which consists of $w^{(l)}_{ji}\in\mathbb{R}$ and $b^{(l)}_j\in\mathbb{R}$ for each $i\in\{1,2,\cdots,n_{l-1}\},\ j\in\{1,2,\cdots,n_l\}$ and $l\in\{1,2,\cdots,L\}$. We define $\bm{\Theta}\subset\mathbb{R}^M$ as the set of $\bm{\theta}$, where $M$ is the number of parameters on the neural network. We call $\bm{g}^{(l)}\circ\bm{f}^{(l)}$ as the $l$-th layer and call the $j$-th coordinate of $\bm{g}^{(l)}\circ \bm{f}^{(l)}$ as the $j$-th unit of the $l$-th layer. We note that the $0$-th layer, which is called as the input layer is not counted into the number of layers.

Deep neural networks are used for supervised learning. We can express the relation between explanatory variables and objective variables with deep neural networks if the values of $\bm{\theta}$ are set appropriately.
\subsection{The Ratio of the Desired Parameters}
\label{secdef1}
In this section, we assume a probability distribution for objective variables in supervised learning and propose a new criterion to evaluate the expressivity of functions computable by deep neural networks under that assumption. For simplicity, we restrict the dimension of objective variables to 1. Let the probability distribution of objective variables be normal distribution such as
\begin{align}
y = F^*(\bm{x})+\varepsilon\ \ (\varepsilon\sim N(0, \sigma^2)), \label{eqnorm}
\end{align}
where $F^* : \mathbb{R}^{n_0}\to\mathbb{R}$ is a continuous function. The goal of deep learning is expressing the relation between explanatory variables and objective variables by approximating $F^*$ with deep neural networks $F_{\bm{\theta}}$ in the set $\{F_{\bm{\theta}}\ |\ \bm{\theta}\in\bm{\Theta}\}$. Therefore, we call $F^*$ as the target function in this paper.

When approximating the target function $F^*$ by deep neural networks $F_{\bm{\theta}}$, it is desirable for the set $\{F_{\bm{\theta}}\ |\ \bm{\theta}\in\bm{\Theta}\}$ to have many functions close to $F^*$. However, there have been few studies which investigate the volume of the functions close to $F^*$ in $\{F_{\bm{\theta}}\ |\ \bm{\theta}\in\bm{\Theta}\}$ although some studies prove that at least one function close to $F^*$ exist in $\{F_{\bm{\theta}}\ |\ \bm{\theta}\in\bm{\Theta}\}$ e.g., \cite{Cybenko, Yarotsky}. Therefore, we define a new criterion to investigate it independently from the efficiency of learning as follows.
\begin{defi}\label{defibetterparam}
Let $F_{\bm{\theta}} : \mathbb{R}^{n_0}\to\mathbb{R}$ be a deep neural network. Let $F^* : \mathbb{R}^{n_0}\to\mathbb{R}$ be the target function defined as (\ref{eqnorm}). Moreover, we assume that explanatory variables $\bm{x}\in\mathbb{R}^{n_0}$ are randam variables with a probability density function $p(\bm{x})$. Then we define \emph{the ratio of the desired parameters} between the set $\mathcal{F}:=\{F_{\bm{\theta}}\ |\ \bm{\theta}\in\bm{\Theta}\}$ and $F^*$ as
\begin{align}
R_{\epsilon}(\mathcal{F}, F^*):=\frac{\int_{\bm{\theta}\in\bm{\Theta}_{\epsilon}(\mathcal{F}, F^*)}d\bm{\theta}}{\int_{\bm{\theta}\in\bm{\Theta}}d\bm{\theta}}, \label{eqR}
\end{align}
where
\begin{align}
d(F_{\bm{\theta}}, F^*)&:=\int_{\bm{x}\in\mathbb{R}^{n_0}}(F_{\bm{\theta}}(\bm{x})-F^*(\bm{x}))^2p(\bm{x})d\bm{x}, \label{eqd}\\
\bm{\Theta}_{\epsilon}(\mathcal{F}, F^*)&:=\{\bm{\theta}\in\bm{\Theta}\ |\ d(F_{\bm{\theta}}, F^*)\leq\epsilon\}. \nonumber
\end{align}
\end{defi}

We can say that $d(F_{\bm{\theta}}, F^*)$ represents the distance between the deep neural network $F_{\bm{\theta}}$ and the target function $F^*$. Moreover, $R_{\epsilon}(\mathcal{F}, F^*)$ calculates the ratio of the values of $\bm{\theta}$ which make that distance short. Therefore, we regard that the larger $R_{\epsilon}(\mathcal{F}, F^*)$ is, the more functions close to the target function the set $\mathcal{F}$ has. 

The ratio of the desired parameters shows the volume of parameter values which give functions close to the target function. Therefore, we can know not only the existence of the appropriate parameter values but also how many the appropriate parameter values exist. If we remind learning with the gradient method, we can consider that the ratio of the desired parameters may show the number of local solutions of the loss function. 
\subsection{Deep Neural Networks and Linear Regions}
\label{secdef}
\subsubsection{ReLU Neural Networks}
We can use some kinds of non-linear functions for the non-linear activation function $\bm{g}^{(l)}$. One example is the ReLU function defined as $g(x)=\max\{0, x\}$. ReLU neural networks have the special property as follows.
\begin{lem}[\cite{Yarotsky}\ ]\label{lem1}
Let $\mathcal{D}$ be a bounded subset of $\mathbb{R}^{n_0}$. Let $\bm{F}_{\bm{\theta}} : \mathcal{D}\to\mathbb{R}^{n_L}$ be a ReLU neural network with $L$ layers, $M$ parameters and $S$ units in total. Then there exists a neural network $\tilde{\bm{F}}_{\bm{\theta}} : \mathcal{D}\to\mathbb{R}^{n_L}$ with any continuous piecewise linear activation function, having $L$ layers, $4M$ parameters and $2S$ units in total, and that computes the same function as $\bm{F}_{\bm{\theta}}$.
\end{lem}

Lemma \ref{lem1} points out that deep neural networks with any piecewise linear activation function can describe any ReLU neural network. In the proof of \cite{Yarotsky}, two units of $\tilde{\bm{F}}_{\bm{\theta}}$ are assigned to the each unit of $\bm{F}_{\bm{\theta}}$ when we replace $\bm{F}_{\bm{\theta}}$ with $\tilde{\bm{F}}_{\bm{\theta}}$. Therefore, the number of units at the $l$-th layer $n_l$ increases to $2n_l$. Lemma \ref{lem1} is used for the proof of the theorem shown in Section \ref{secbene}.
\begin{figure}[t]
\begin{center}
\centerline{\includegraphics[height=4.5cm,width=7cm]{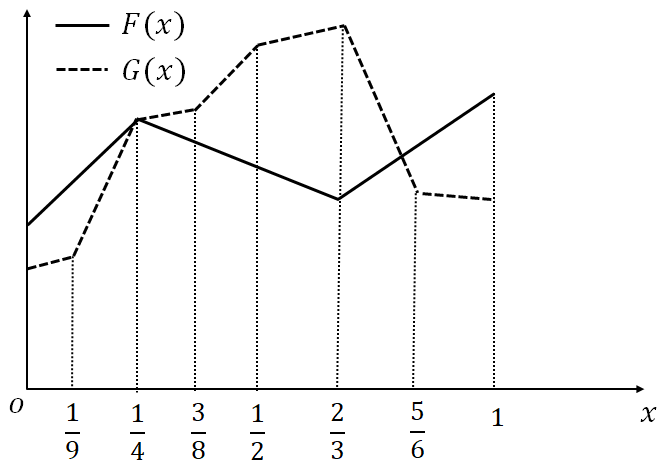}}
\caption{The examples of linear regions of piecewise linear functions.}
\label{fig5}
\end{center}
\end{figure}
\subsubsection{Fineness of Linear Regions}
The ReLU function $g(x)=\max\{0, x\}$ is a piecewise linear function whose inclination changes at the origin. If a neural network has piecewise linear activation functions like this, $\bm{f}^{(l)}(l=1,2,\cdots,L)$ are linear functions and $\bm{g}^{(l)}(l=1,2,\cdots,L-1)$ are piecewise linear functions. Therefore, a neural network whose activation functions are piecewise linear is a piecewise linear function. Based on this fact, we can introduce a \emph{linear region} to consider the flexibility of deep neural networks with piecewise linear activation functions. It is defined as follows. 
\begin{defi}[\cite{Montufar}\ ]\label{defi1}
Let $\mathcal{D}$ be a bounded subset of $\mathbb{R}^{n}$. Let $\bm{F} : \mathcal{D}\to\mathbb{R}^{m}$ be a piecewise linear function. Let $U$ be a $n$-dimensional interval in $\mathcal{D}$. Let $\bm{F}|_U : U\to\mathbb{R}^{m}$ be the function given by restricting the input domain of $\bm{F}$ from $\mathcal{D}$ to $U$. We say that $U$ is a \emph{linear region} of $\bm{F}$ if the followings hold. 
\begin{itemize}
\item $\bm{F}|_U$ is linear. 
\item For any subset $V\subset\mathcal{D}$ such that $V\supsetneq U$, $\bm{F}|_V$\ is not linear.
\end{itemize}
\end{defi}
\begin{ex}
For example, let $F : [0,1]\to\mathbb{R}$ be the function in Fig. \ref{fig5}. We can say that $[0, \frac{1}{4}]$, $[\frac{1}{4}, \frac{2}{3}]$ and $[\frac{2}{3}, 1]$ are linear regions of $F$. $[\frac{1}{3}, \frac{1}{2}]$ is not a linear region. 
\end{ex}
\begin{figure}[t]
\begin{center}
\centerline{\includegraphics[height=3.7cm,width=9.5cm]{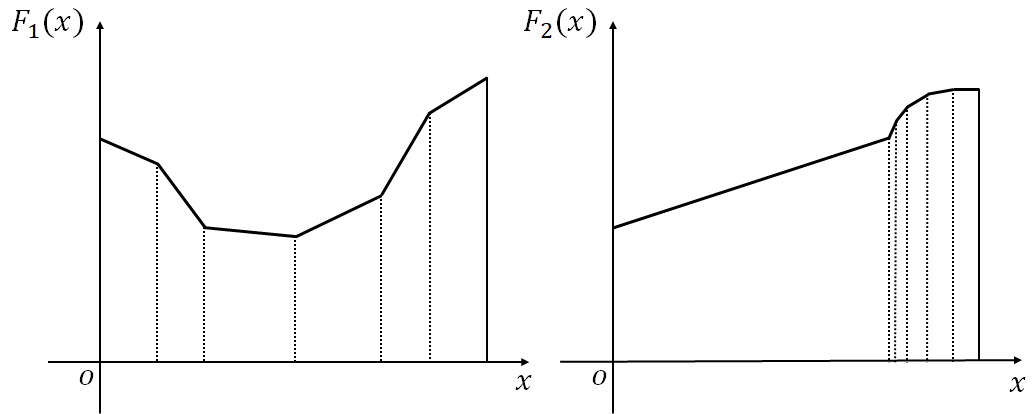}}
\caption{The problem of evaluating the flexibility of deep neural networks by the number of linear regions. We cannot distinguish between $F_1$ and $F_2$ when we use the number of linear regions. Therefore, we propose a new criterion in this paper.}
\label{fig7}
\end{center}
\end{figure}

In this paper, the set of linear regions of $\bm{F}$ is represented by $L(\bm{F})$. We note that 
\begin{align}
\bigcup_{U\in L(\bm{F})}U=\mathcal{D},\ \ \ \ \ \sum_{U\in L(\bm{F})}\int_{\bm{x}\in U}d\bm{x}=\int_{\bm{x}\in\mathcal{D}}d\bm{x} \label{eqpiece}
\end{align}
when $\bm{F} : \mathcal{D}\to\mathbb{R}^{m}$ is a piecewise linear function.

There have been some studies that evaluate the flexibility of deep neural networks by the number of linear regions \cite{Pascanu, Montufar, Serra}. However, there is a room for improvement in the evaluation of the flexibility by the number of linear regions when we want to know whether linear regions are scattered uniformly. The number of linear regions cannot ensure that many linear regions do not gather in a certain area in the input domain. This problem appears in Fig. \ref{fig7}. $F_2$ does not keep its flexibility in the left-side area. However, $F_1$ and $F_2$ are regarded that they have similar flexibility by the number of linear regions. There is also the same problem in the evaluation of the flexibility by the volume of the boundary between two linear regions proposed in \cite{Hanin}. Therefore, we propose a new criterion to evaluate the flexibility of deep neural networks. It is defined as follows.
\begin{defi}\label{defi2}
Let $\mathcal{D}$ be a bounded subset of $\mathbb{R}^{n}$. We define the \emph{fineness} of linear regions of a piecewise linear function $\bm{F} : \mathcal{D}\to\mathbb{R}^{m}$ such as
\begin{align}
I(\bm{F}):=\max_{U\in L(\bm{F})}\frac{\int_{\bm{x}\in U}d\bm{x}}{\int_{\bm{x}\in \mathcal{D}}d\bm{x}}.
\label{eqif}
\end{align}
\end{defi}
\begin{ex}
We consider the example function $F : [0,1]\to\mathbb{R}$ in Fig. \ref{fig5}. In this example, the fineness of linear regions $I(F)$ is computed as follows.
\begin{align}
I(F)&=\max_{U\in\left\{\left[0, \frac{1}{4}\right], \left[\frac{1}{4}, \frac{2}{3}\right], \left[\frac{2}{3}, 1\right]\right\}}\frac{\int_{x\in U}dx}{\int_{x\in [0,1]}dx}=\int^{\frac{2}{3}}_{\frac{1}{4}}dx=\frac{5}{12}. \nonumber
\end{align}
\end{ex}

Note that $0\leq I(\bm{F})\leq 1$. We regard that the smaller the value of $I(\bm{F})$ is, the finer the linear regions of $\bm{F}$ are. The fineness of linear regions shows the size of the maximal linear region of piecewise linear functions and shows how fine linear regions the area which is the least flexible has. Therefore, we can know whether linear regions are scattered finely and uniformly on the input domain or not, by that criterion. It is different from the number of linear regions. In the case of neural networks $\bm{F}_{\bm{\theta}}$ whose activation functions are piecewise linear, the fineness of linear regions $I(\bm{F}_{\bm{\theta}})$ changes when the values of $\bm{\theta}$ change. Therefore, we regard that the smaller the value of $\min_{\bm{\theta}\in\bm{\Theta}}I(\bm{F}_{\bm{\theta}})$ is, the more flexibility the set $\{\bm{F}_{\bm{\theta}}\ |\ \bm{\theta}\in\bm{\Theta}\}$ has. 

For the readers who are interested in the relation between the number of linear regions and the fineness of linear regions, we show the following fact.
\begin{coro}\label{coro1}
Let $\mathcal{D}$ be a bounded subset of $\mathbb{R}^{n}$. Let $\bm{F} : \mathcal{D}\to\mathbb{R}^{m}$ be a piecewise linear function. The fineness of linear regions $I(\bm{F})$ and the number of linear regions $|L(\bm{F})|$ satisfy 
\begin{align}
|L(\bm{F})|\geq \frac{1}{I(\bm{F})}. \nonumber
\end{align}
\end{coro}
\begin{proofc}
From (\ref{eqpiece}),
\begin{align}
\int_{\bm{x}\in\mathcal{D}}d\bm{x}&=\sum_{U\in L(\bm{F})}\int_{\bm{x}\in U}d\bm{x} \nonumber\\
&\leq |L(\bm{F})|\times \max_{U\in L(\bm{F})}\int_{\bm{x}\in U}
d\bm{x}. \nonumber
\end{align}
Therefore, 
\begin{align}
|L(\bm{F})|\geq \frac{\int_{\bm{x}\in\mathcal{D}}d\bm{x}}{\max_{U\in L(\bm{F})}\int_{\bm{x}\in U}d\bm{x}}=\frac{1}{I(\bm{F})}\ \ \ \ (\because\ (\ref{eqif})). \nonumber
\end{align}
\qed
\end{proofc}

\subsubsection{The Partial Order of Piecewise Linear Functions}
Next, we define a partial order between two piecewise linear functions as follows.
\begin{defi}\label{defi3}
Let $\mathcal{D}$ be a bounded subset of $\mathbb{R}^{n}$. Let $\bm{F}, \bm{G} : \mathcal{D}\to\mathbb{R}^{m}$ be two piecewise linear functions. We say that $\bm{G}\preceq_r \bm{F}$, where $0\leq r\leq 1$, if the following statements hold.
\begin{itemize}
\item For any linear region $U\in L(\bm{F})$, there exist $s\in\mathbb{N}\ (s\geq1)$ and some linear regions $V_i\in L(\bm{G})\ (i=1,2,\cdots,s)$ such that
\begin{align}
U=\bigcup^{s}_{i=1}V_i. \label{equ}
\end{align}
\item For any linear region $U\in L(\bm{F})$ and $V\in\{V'\in L(\bm{G})\ |\ V'\subset U\}$,
\begin{align}
\int_{\bm{x}\in V}d\bm{x}\leq r\int_{\bm{x}\in U}d\bm{x}. \label{eqr}
\end{align}
\end{itemize}
\end{defi}
\begin{ex}
For example, let $F, G : [0,1]\to\mathbb{R}$ be the functions in Fig. \ref{fig5}. We can say that $G\preceq_{\frac{5}{9}} F$ because $r$ is computed such as 
\begin{align}
r=\frac{\frac{1}{4}-\frac{1}{9}}{\frac{1}{4}-0}=\frac{5}{9}. \nonumber
\end{align}
\end{ex}

Equation (\ref{equ}) claims that the linear regions of $\bm{G}$ are finer than those of $\bm{F}$ and (\ref{eqr}) claims that the linear regions of $\bm{G}$ are fine $r$ times or more those of $\bm{F}$. In fact, we can say the following. 
\begin{lem}\label{lem4}
Let $\mathcal{D}$ be a bounded subset of $\mathbb{R}^{n}$. Let $\bm{F}, \bm{G} : \mathcal{D}\to\mathbb{R}^{m}$ be two piecewise linear functions. We can say that
\begin{align}
\bm{G}\preceq_r \bm{F}\ \Rightarrow\ I(\bm{G})\leq rI(\bm{F}), \nonumber
\end{align}
where $0\leq r\leq 1$.
\end{lem}
\setcounter{proofl}{1}
\begin{proofl}
\begin{align}
I(\bm{G})&=\max_{V\in L(\bm{G})}\frac{\int_{\bm{x}\in V}d\bm{x}}{\int_{\bm{x}\in\mathcal{D}}d\bm{x}}\ \ \ \ (\because (\ref{eqif})) \nonumber\\
&=\max_{U\in L(\bm{F})}\max_{V\in\{V'\in L(\bm{G})\ |\ V'\subset U\}}\frac{\int_{\bm{x}\in V}d\bm{x}}{\int_{\bm{x}\in\mathcal{D}}d\bm{x}}\ \ \ \ (\because (\ref{equ})) \nonumber\\
&\leq \max_{U\in L(\bm{F})}r\frac{\int_{\bm{x}\in U}d\bm{x}}{\int_{\bm{x}\in\mathcal{D}}d\bm{x}}\ \ \ \ (\because (\ref{eqr}))\nonumber\\
&=rI(\bm{F}) \ \ \ \ (\because (\ref{eqif})).\nonumber
\end{align}
\qed
\end{proofl}

Lemma \ref{lem4} will be used for the proof of the theorem shown in Section \ref{secbene}.
\subsubsection{Linear Regions and Identification of Input Subsets}
Montufar et al. \cite{Montufar} define the identification of input subsets in their paper. It is defined as follows.
\begin{defi}[\cite{Montufar}\ ]\label{defi4}
We say that a map $\bm{g} : \mathbb{R}^n\to\mathbb{R}^{m}$ \emph{identifies} $K(\in\mathbb{N})$ subsets $U_1,\cdots,U_K(\subset\mathbb{R}^n)$ onto $V(\subset\mathbb{R}^m)$ if it maps them to a commom image $V=\bm{g}(U_1)=\cdots=\bm{g}(U_K)$ in $\mathbb{R}^{m}$. We also say that $U_1,\cdots,U_K$ are \emph{identified} onto $V$ by $\bm{g}$. 
\end{defi}
\begin{ex}
For example, the four quadrants of two dimensional euclidean space are identified  onto $[0,\infty)^2$ by the function $\bm{g} : \mathbb{R}^2\to[0, \infty)^2$ such as $\bm{g}(x_1, x_2)=(|x_1|, |x_2|)^T$.
\end{ex}

If $U_1,\cdots,U_K(\subset\mathbb{R}^n)$ are all of the subsets identified onto $V(\subset\mathbb{R}^{m})$ by $\bm{g} : \mathbb{R}^n\to\mathbb{R}^{m}$, we can say that
\begin{align}
\bm{g}^{-1}(V)=\bigcup^K_{k=1}U_k. \label{eqid}
\end{align}
Montufar et al. \cite{Montufar} interpret (\ref{eqid}) that the common subset $V$ is replicated to $U_1,\cdots,U_K$. 
We can prove the following lemma related to Definition \ref{defi4}.
\begin{lem}\label{lem3}
Let $\mathcal{D}_1$ be a bounded subset of $\mathbb{R}^{n}$. Let $\mathcal{D}_2$ be a bounded subset of $\mathbb{R}^{n'}$. Let $\bm{g} : \mathcal{D}_1\to\mathcal{D}_2$ be a piecewise linear function identifying all its linear regions onto $\mathcal{D}_2$. Let $\bm{f} : \mathcal{D}_2\to\mathbb{R}^{m}$ be a piecewise linear function. The following holds.
\begin{align}
\bm{f}\circ \bm{g}\preceq_{I(\bm{f})}\bm{g}. \nonumber
\end{align}
\end{lem}
\begin{proofl}
For any linear region $U\in L(\bm{g})$ and $V\in L(\bm{f})$,\ $\bm{g}|^{-1}_U(V)$ is a linear region of $\bm{f} \circ \bm{g}$ and 
\begin{align}
U&=\bm{g}|_U^{-1}(\mathcal{D}_2)\ \ \ \ (\because \bm{g}(U)=\mathcal{D}_2)\nonumber\\
&=\bm{g}|_U^{-1}\left(\bigcup_{V\in L(\bm{f})}V\right)\ \ \ \ (\because (\ref{eqpiece})) \nonumber\\
&=\bigcup_{V\in L(\bm{f})}\bm{g}|_U^{-1}(V). \nonumber
\end{align}

Since $\bm{g}|_U^{-1}$ is linear, the jacobian of $\bm{g}|_U^{-1}$ satisfies $\left|J_{\bm{g}|_U^{-1}}\right|\neq0$. Therefore, from the property of integration by substitution with $\bm{x}=\bm{g}|_U^{-1}(\bm{y})$,
\begin{align}
\int_{\bm{x}\in U}d\bm{x}&=\int_{\bm{x}\in \bm{g}|_U^{-1}(\mathcal{D}_2)}d\bm{x}=\left|J_{\bm{g}|_U^{-1}}\right|\int_{\bm{y}\in\mathcal{D}_2}d\bm{y}. \label{eqk}
\end{align}
From (\ref{eqk}), 
\begin{align}
\left|J_{\bm{g}|_U^{-1}}\right|=\frac{\int_{\bm{x}\in U}d\bm{x}}{\int_{\bm{y}\in\mathcal{D}_2}d\bm{y}}. \label{eqj}
\end{align}
Moreover,
\begin{align}
\int_{\bm{x}\in\bm{g}|_U^{-1}(V)}d\bm{x}&=\left|J_{\bm{g}|_U^{-1}}\right|\int_{\bm{y}\in V}d\bm{y} \nonumber\\
&=\frac{\int_{\bm{x}\in U}d\bm{x}}{\int_{\bm{y}\in\mathcal{D}_2}d\bm{y}}\int_{\bm{y}\in V}d\bm{y}\ \ \ \ (\because \mbox{(\ref{eqj})}) \nonumber\\
&\leq I(\bm{f})\int_{\bm{x}\in U}d\bm{x}\ \ \ \ (\because (\ref{eqif})). \nonumber
\end{align}
\qed
\end{proofl}
\subsubsection{The Relation Between the Depth of Neural Networks and the Fineness of Linear Regions}
\label{secbene}
We can show that to increase layers is more effective than to increase units at each layer on improving the expressivity of neural networks, by proving the following theorem.
\begin{theo}\label{theo}
Let $\tilde{\bm{F}}_{\bm{\theta}} : [0,1]^{n_0}\to\mathbb{R}^{n_L}$ be a neural network 
with any piecewise linear activation function, having $L$ layers and $n_l$ units at the $l$-th layer, where $n_l\geq 2n_0$ for any $l\in\{1,2,\cdots,L-1\}$. We can say that
\begin{align}
^\exists\bm{\theta}\in\bm{\Theta}\ \ \mathrm{s.t.}\ \ I(\tilde{\bm{F}}_{\bm{\theta}})\leq \prod^{L-1}_{l=1}\left\lfloor\frac{n_l}{2n_0}\right\rfloor^{-n_0}. \label{eqtheo}
\end{align}
\end{theo}

If $n_l=n(\geq2n_0)$ for any $l\in\{1,2,\cdots,L-1\}$, we can say that
\begin{align}
^\exists\bm{\theta}\in\bm{\Theta}\ \ \mathrm{s.t.}\ \ I(\tilde{\bm{F}}_{\bm{\theta}})\leq O\left(\left(\frac{n}{2n_0}\right)^{-n_0(L-1)}\right) \nonumber
\end{align}
from Theorem \ref{theo}. In other words, $\tilde{\bm{F}}_{\bm{\theta}}$ can express functions which have the fineness of linear regions $O((\frac{n}{2n_0})^{-n_0(L-1)})$ if $\bm{\theta}$ has appropriate values. Therefore, we consider that the flexibility of the set of deep neural networks $\{\tilde{\bm{F}}_{\bm{\theta}}\ |\ \bm{\theta}\in\bm{\Theta}\}$ grows exponentially in $L$ and polynominally in $n$. 
We use the following lemma in order to prove Theorem \ref{theo}.
\begin{lem}\label{lem2}
Let $\bm{F}_{\bm{\theta}} : [0,1]^{n_0}\to\mathbb{R}^{n_L}$ be a ReLU neural network with $L$ layers and $n_l$ units at the $l$-th layer, where $n_l\geq n_0$ for any $l\in\{1,2,\cdots,L-1\}$. We can say that
\begin{align}
^\exists{\bm{\theta}}\in\bm{\Theta}\ \ \mathrm{s.t.}\ \ I(\bm{F}_{\bm{\theta}})\leq \prod^{L-1}_{l=1}\left\lfloor\frac{n_l}{n_0}\right\rfloor^{-n_0}. \nonumber
\end{align}
\end{lem}
\setcounter{proofl}{3}
\begin{prooft}
Let $\bm{F}_{\bm{\theta}} : [0,1]^{n_0}\to\mathbb{R}^{n_L}$ be a ReLU neural network with $L$ layers and $\frac{n_l}{2}(n_l\geq 2n_0)$ units at the $l$-th layer for any $l\in\{1,2,\cdots,L-1\}$. From Lemma \ref{lem2}, there exists $\hat{\bm{\theta}}\in\bm{\Theta}$ such that
\begin{align}
I(\bm{F}_{\hat{\bm{\theta}}})\leq \prod^{L-1}_{l=1}\left\lfloor\frac{n_l}{2n_0}\right\rfloor^{-n_0}. \nonumber
\end{align}

Moreover, from Lemma \ref{lem1}, there exists a neural network $\tilde{\bm{F}}_{\bm{\theta}} : [0,1]^{n_0}\to\mathbb{R}^{n_L}$ with any piecewise linear activation function, having $L$ layers and $n_l$ units at the $l$-th layer, and that computes the same function as $\bm{F}_{\hat{\bm{\theta}}}$. It satisfies 
\begin{align}
I(\tilde{\bm{F}}_{\bm{\theta}})\leq \prod^{L-1}_{l=1}\left\lfloor\frac{n_l}{2n_0}\right\rfloor^{-n_0}. \nonumber
\end{align}
\qed
\end{prooft}

Now, we prove Lemma \ref{lem2}. 
\begin{proofl}
We prove Lemma \ref{lem2} by two steps as follows. 
\begin{description}
\item[Step 1.]\ Construct a specific ReLU neural network $\bm{F}_{\hat{\bm{\theta}}}$ which recursively caluculates the functions with common parameters.
\item[Step 2.]\ Show that the fineness of linear regions of the network $\bm{F}_{\hat{\bm{\theta}}}$ satisfies $I(\bm{F}_{\hat{\bm{\theta}}})\leq\prod^{L-1}_{l=1}\lfloor\frac{n_l}{n_0}\rfloor^{-n_0}$ with applying Lemma \ref{lem4} and Lemma \ref{lem3} inductively.
\end{description}

\textbf{\underline{Step 1}.} Let an integer $n\in\mathbb{N}$ satisfy $n\geq n_0$ and $p:=\lfloor\frac{n}{n_0}\rfloor$. We define a $pn_0$-dimensional function $\tilde{\bm{f}} : [0,1]^{n_0}\to\mathbb{R}^{pn_0}$ as 
\begin{align}
&\tilde{f}_{(i-1)n_0+j}(\bm{x}):=\left\{\begin{array}{ll}
px_j&(i=1)\\
2px_j-2(i-1)&(i=2,\cdots,p)
\end{array}\right. \nonumber \\
&\hspace{5cm} (j=1,2,\cdots,n_0), \nonumber
\end{align}
where $\tilde{f}_{(i-1)n_0+j}$ is the $((i-1)n_0+j)$-th coordinate of $\tilde{\bm{f}}$ and $x_j$ is the $j$-th coordinate of $\bm{x}$. We define a $n_0$-dimensional function $\tilde{\tilde{\bm{f}}} : \mathbb{R}^{pn_0}\to[0,1]^{n_0}$ as
\begin{align}
\tilde{\tilde{f}}_j(\bm{x}):=\sum^p_{i=1}(-1)^{i-1}x_{(i-1)n_0+j}\ (j=1,2,\cdots,n_0), \nonumber
\end{align}
where $\tilde{\tilde{f}}_j$ is the $j$-th coordinate of $\tilde{\tilde{\bm{f}}}$. Furthermore, we define a $n_L$-dimensional function $\tilde{\tilde{\tilde{\bm{f}}}} : [0,1]^{n_0}\to\mathbb{R}^{n_L}$ as
\begin{align}
\tilde{\tilde{\tilde{f}}}_j(\bm{x}):=\sum^{n_0}_{i=1}w_{ji}x_i+b_j\ (w_{ji}>0;\ j=1,2,\cdots,n_L), \nonumber
\end{align}
where $\tilde{\tilde{\tilde{f}}}_j$ is the $j$-th coordinate of $\tilde{\tilde{\tilde{\bm{f}}}}$. We construct the neural network using these functions as follows.

At first, let the number of units at the $l$-th layer be equal to $n(\geq n_0)$ for any $l\in\{1,2,\cdots,L-1\}$. We separate the set of $n$ units at the $l$-th layer, $(l=1,2,\cdots,L-1)$, into $p(=\lfloor\frac{n}{n_0}\rfloor)$ subsets and remainder units. Each of the $p$ subsets contains $n_0$ units. Let the values of the parameters of the remainder $n-pn_0$ units be $0$. For any $l\in\{1,2,\cdots,L-1\}$, let the non-linear activation function $\bm{g}^{(l)} : \mathbb{R}^{pn_0}\to\mathbb{R}^{pn_0}$ be 
\begin{align}
g^{(l)}_j(\bm{x})=\max\{0, x_j\}\ (j=1,2,\cdots,pn_0), \nonumber
\end{align}
where $g^{(l)}_j$ is the $j$-th coordinate of $\bm{g}^{(l)}$.

Next, we consider the linear functions $\bm{f}^{(l)}(l=1,2,\cdots,L)$. When $l=1$, for the $j$-th unit of the $i$-th subset, ($j = 1, 2, \dots , n_0$, $i = 1, 2, \dots , p$), let the linear function $f^{(1)}_{(i-1)n_0+j} : [0,1]^{n_0}\to\mathbb{R}$ be
\begin{align}
f^{(1)}_{(i-1)n_0+j}=\tilde{f}_{(i-1)n_0+j}. \nonumber
\end{align}
When $l=2, 3, \dots , L-1$, for the $j$-th unit of the $i$-th subset, ($j = 1, 2, \dots , n_0$, $i = 1, 2, \dots , p$), let the linear function $f^{(l)}_{(i-1)n_0 + j} : \mathbb{R}^{pn_0}\to\mathbb{R}$ be 
\begin{align}
f_{(i-1)n_0 + j}^{(l)}=\tilde{f}_{(i-1)n_0 + j}\circ \tilde{\tilde{\bm{f}}}. \nonumber
\end{align}
When $l=L$, let the linear function $\bm{f}^{(L)} : \mathbb{R}^{pn_0}\to\mathbb{R}^{n_L}$ be 
\begin{align}
\bm{f}^{(L)}=\tilde{\tilde{\tilde{\bm{f}}}}\circ \tilde{\tilde{\bm{f}}}.\nonumber
\end{align}

Then the constructed network is represented as follows. Since $\bm{g}^{(1)}, \bm{g}^{(2)}, \cdots, \bm{g}^{(L-1)}$ are common, we express them as $\bm{g}$.
\begin{align}
\bm{F}_{\hat{\bm{\theta}}}&=\bm{f}^{(L)}\circ \bm{g}^{(L-1)}\circ \bm{f}^{(L-1)}\circ\cdots\circ \bm{g}^{(1)}\circ \bm{f}^{(1)} \nonumber \\
&=\tilde{\tilde{\tilde{\bm{f}}}} \circ \tilde{\tilde{\bm{f}}} \circ \bm{g} \circ \tilde{\bm{f}} \circ \cdots \circ \tilde{\tilde{\bm{f}}} \circ \bm{g} \circ \tilde{\bm{f}} \circ \tilde{\tilde{\bm{f}}} \circ \bm{g} \circ \tilde{\bm{f}}. \nonumber
\end{align}
\begin{figure}[t]
\begin{center}
\centerline{\includegraphics[height=5cm,width=9cm]{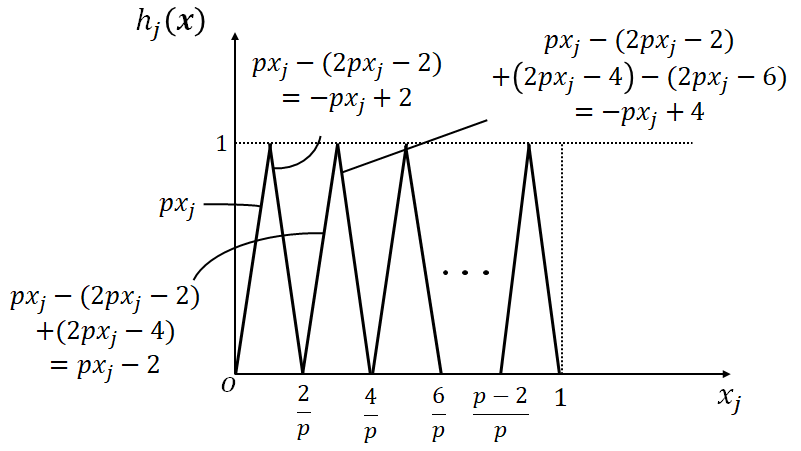}}
\caption{The graph of $h_j\ (j=1,\cdots,n_0)$, where $p$ is a even number.}
\label{fig1}
\end{center}
\end{figure}

\textbf{\underline{Step 2}.} Let $\bm{h} := \tilde{\tilde{\bm{f}}} \circ \bm{g} \circ \tilde{\bm{f}}$. We can re-represent our network as
\begin{align}
\bm{F}_{\hat{\bm{\theta}}}=\tilde{\tilde{\tilde{\bm{f}}}} \circ \underbrace{\bm{h} \circ \bm{h} \circ \cdots \circ \bm{h}}_{L-1}. \nonumber
\end{align}
We consider the fineness of linear regions $I(\tilde{\tilde{\tilde{\bm{f}}}})$ and $I(\bm{h})$ as well as the identified regions of $\bm{h}$ in order to apply Lemma \ref{lem4} and Lemma \ref{lem3}.  Since $\tilde{\tilde{\tilde{\bm{f}}}}$ is linear on $[0,1]^{n_0}$,
\begin{align}
I(\tilde{\tilde{\tilde{\bm{f}}}})=1. \nonumber
\end{align}
On the other hand, the $j$-th coordinate of $\bm{h} : [0,1]^{n_0}\to[0,1]^{n_0}$ is described such as
\begin{align}
h_j(\bm{x})&=\max\left\{0,\ px_{j}\right\}-\max\left\{0,\ 2px_{j}-2\right\}+\cdots \nonumber\\
&\hspace{1cm}\cdots+(-1)^{p-1}\max\left\{0,\ 2px_{j}-2(p-1)\right\} \nonumber\\
&\hspace{4cm}(j=1,2,\cdots,n_0). \nonumber
\end{align}
Figure \ref{fig1} shows the graph of $h_j$. From Fig. \ref{fig1}, we can say that $h_j$ identifies its linear regions 
\begin{align}
\left[\frac{t}{p}, \frac{t+1}{p}\right]\ (t=0,1,\cdots,p-1) \nonumber
\end{align}
onto $[0,1]$. In other words, $\bm{h}$ identifies these linear regions on the each coordinate $h_1,h_2,\cdots,h_{n_0}$. Therefore, it identifies its linear regions 
\begin{align}
\prod^{n_0}_{j=1}\left[\frac{t_j}{p}, \frac{t_j+1}{p}\right]\ (t_j=0,1,\cdots,p-1) \nonumber
\end{align}
onto $[0,1]^{n_0}$. Since these linear regions are $n_0$-dimensional hypercubes with the side length of $p^{-1}$ for any $t_j\in\{0,1,\cdots,p-1\}$, the fineness of linear regions $I(\bm h)$ satisfies 
\begin{align}
I(\bm{h})=p^{-n_0}. \nonumber
\end{align}

Then we can say that 
\begin{align}
\tilde{\tilde{\tilde{\bm f}}}\circ \bm{h}\preceq_{I(\tilde{\tilde{\tilde{\bm{f}}}})}\bm{h} \nonumber
\end{align}
from Lemma \ref{lem3}. Therefore, from Lemma \ref{lem4}, 
\begin{align}
I(\tilde{\tilde{\tilde{\bm{f}}}}\circ \bm{h})\leq I(\tilde{\tilde{\tilde{\bm{f}}}})I(\bm{h})=p^{-n_0}. \label{eqiii}
\end{align}
In a similar manner, from Lemma \ref{lem3}, 
\begin{align}
(\tilde{\tilde{\tilde{\bm{f}}}}\circ \bm{h})\circ \bm{h}\preceq_{I(\tilde{\tilde{\tilde{\bm{f}}}}\circ \bm{h})}\bm{h}. \nonumber
\end{align}
Therefore, from Lemma \ref{lem4} and (\ref{eqiii}),
\begin{align}
I(\tilde{\tilde{\tilde{\bm{f}}}}\circ \bm{h}\circ \bm{h})\leq I(\tilde{\tilde{\tilde{\bm{f}}}}\circ \bm{h})I(\bm{h})
\leq I(\tilde{\tilde{\tilde{\bm{f}}}})I(\bm{h})I(\bm{h})
=p^{-2n_0}. \nonumber
\end{align}

If we inductively repeat these operations, we get the inequation as follows.
\begin{align}
I(\bm{F}_{\hat{\bm{\theta}}})&=I(\tilde{\tilde{\tilde{\bm{f}}}} \circ \underbrace{\bm{h} \circ \bm{h} \circ \cdots \circ \bm{h}}_{L-1}) \nonumber\\
&\leq I(\tilde{\tilde{\tilde{\bm{f}}}})\underbrace{I(\bm{h})I(\bm{h})\cdots I(\bm{h})}_{L-1}=p^{-n_0(L-1)}. \nonumber
\end{align}

If the number of units at each layer is different from each other and satisfies $n_l\geq n_0(l=1,2,\cdots,L-1)$, we may change $p=\lfloor\frac{n}{n_0}\rfloor$ to 
\begin{align}
p_l:=\left\lfloor\frac{n_l}{n_0}\right\rfloor\ (l=1,2,\cdots,L-1).  \nonumber
\end{align}
In this case, the fineness of linear regions $I(\bm{F}_{\hat{\bm{\theta}}})$ satisfies 
\begin{align}
I(\bm{F}_{\hat{\bm{\theta}}})\leq\prod^{L-1}_{l=1}p_l^{-n_0}. \nonumber
\end{align}
\qed
\end{proofl}
\section{Experiments}
\begin{table}[t]
\caption{The neural networks compared in the computer simulations.}
\label{tabmodel}
\centering
\begin{tabular}{|c|cc|}\hline
$l$&Network 1 ($L=6$)&Network 2 ($L=2$)\\\hline
0&$n_0=1$&$n_0=1$\\
1&$n_1=4$&$n_1=20$\\
2&$n_2=4$&$n_2=1$\\
3&$n_3=4$&\\
4&$n_4=4$&\\
5&$n_5=4$&\\
6&$n_6=1$&\\\hline
\end{tabular}
\end{table}

We compared two types of neural networks shown in Table \ref{tabmodel} in terms of the expressivity. The activation functions in those neural networks are the ReLU function. Although the depth of them is different, the number of units is equal to 22. We approximately calculated the ratio of the desired parameters and the fineness of linear regions by computer simulations in order to evaluate the expressivity of the two neural networks.
\subsection{The Calculation of the Ratio of the Desired Parameters}
\label{secexex}

\begin{figure}[t]
\begin{center}
\centerline{\includegraphics[height=4.5cm,width=7.5cm]{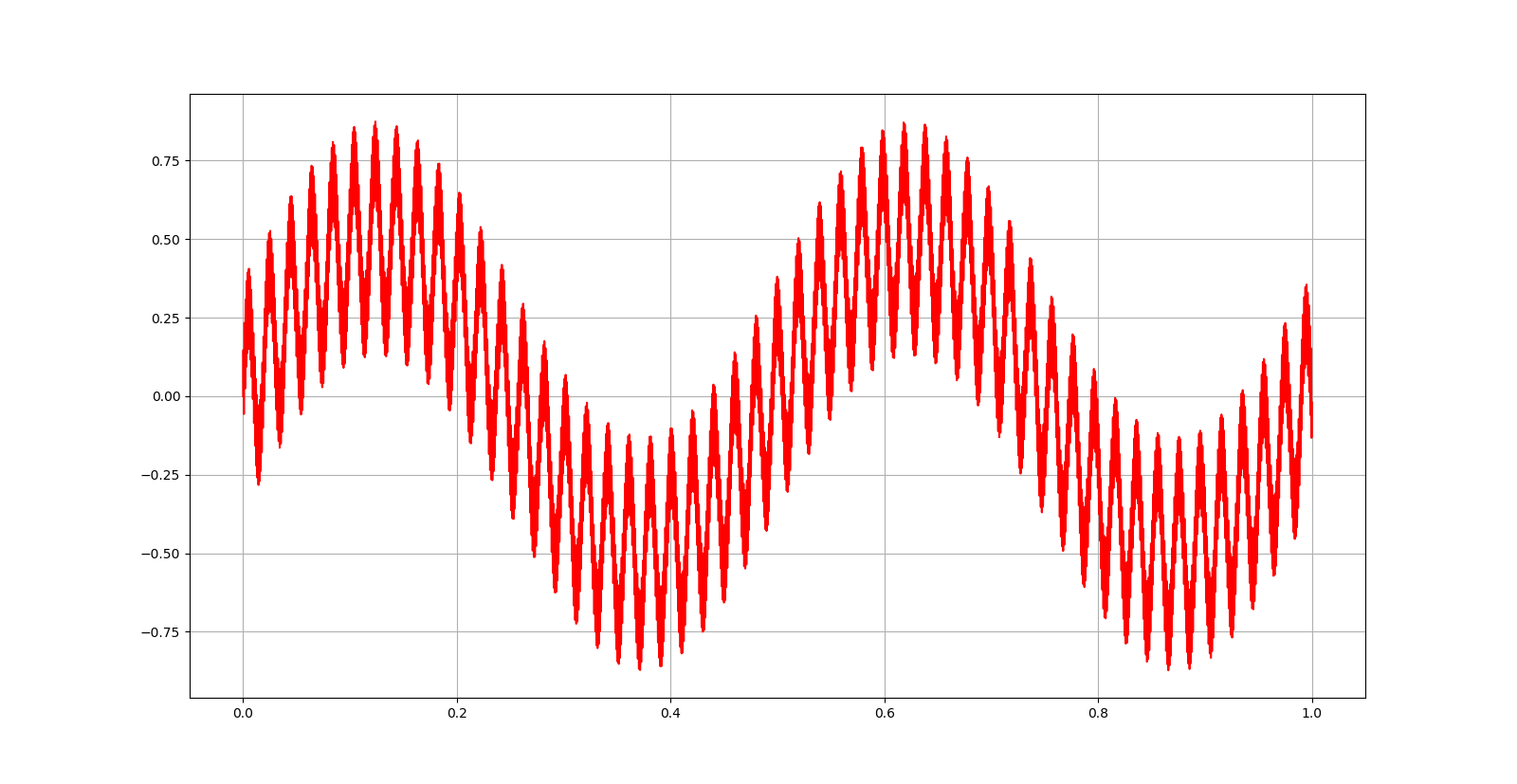}}
\caption{The Weierstrass function we used in the computer simulations.}
\label{figweier}
\end{center}
\end{figure}

We calculated the ratio of the desired parameters by computer simulations. We executed the calculation with restricting the input and output dimension to one for simplicity.  It is difficult to calculate (\ref{eqd}) by computer simulations. Therefore, instead of it, we calculated such as
\begin{align}
\hat{d}(F_{\bm{\theta}}, F^*):=\frac{1}{N}\sum^N_{n=1}(F_{\bm{\theta}}(x_n)-F^*(x_n))^2, \nonumber
\end{align}
where $x_n\in\mathbb{R}(n=1,2,\cdots,N)$ are finite samples. For the target function $F^*$, we adopted the two kinds of functions. One was the sin function $F^*(x)=\sin(4\pi x)$ and the other was the Weierstrass function such as Fig. \ref{figweier}. The samples $x_n(n=1,2,\cdots,N)$ were set as $x_n=(n-1)\times 10^{-4}$ and $N=10^4$. 

Moreover, it is also difficult to calculate (\ref{eqR}) by computer simulations. Therefore, we approximately calculated it with random sampling of the values of $\bm{\theta}$. We generated the values of $\bm{\theta}$ with random sampling of standard normal distribution and calculated $\hat{d}(F_{\bm{\theta}}, F^*)$ with the generated values of $\bm{\theta}$, $2\times 10^4$ times. Then we confirmed whether the each $\hat{d}(F_{\bm{\theta}}, F^*)$ was less than $\epsilon_k(k=1,2,\cdots,10^4)$ or not. When the target function was $\sin(4\pi x)$, $\epsilon_k=0.4+4(k-1)\times10^{-5}$. When the target function was the Weierstrass function, $\epsilon_k=0.6+2(k-1)\times10^{-5}$. We counted up when $\hat{d}(F_{\bm{\theta}}, F^*)\leq\epsilon_k$ and calculated the ratio of $\bm{\theta}$ which satisfied the inequation by dividing the last count by $2\times 10^4$. The following summarizes the algorithm which approximately calculates $R_{\epsilon}(\mathcal{F}, F^*)$.
\begin{algorithmic}
\STATE $\epsilon_k\ \leftarrow\ 0.4+4(k-1)\times10^{-5}$ or $0.6+2(k-1)\times10^{-5}\ (k=1,2,\cdots,10^4)$
\STATE $t_{k}\ \leftarrow\ 0\ (k=1,2,\cdots,10^4)$
\STATE $x_n\ \leftarrow\ (n-1)\times 10^{-4}\ (n=1,2,\cdots,10^4)$
\STATE $F^*(x_n)\ \leftarrow\ \sin(4\pi x_n)$ or the Weierstrass function\ $(n=1,2,\cdots,10^4)$
\FOR{$j=1$ {\bfseries to} $2\times 10^4$}
	\STATE Generate $\bm{\theta}$ with random sampling of standard normal distribution
	\STATE Compute a neural network $F_{\bm{\theta}}(x_n)\ (n=1, 2, \cdots, 10^4)$ 
	\STATE $\hat{\mu}\ \leftarrow\ \frac{1}{10^4}\sum^{10^4}_{n=1}F_{\bm{\theta}}(x_n)$
	\STATE $\hat{\sigma}\ \leftarrow\ \sqrt{\frac{1}{10^4}\sum^{10^4}_{n=1}\left(F_{\bm{\theta}}(x_n)-\hat{\mu}\right)^2}$
	\STATE $F_{\bm{\theta}}(x_n)\ \leftarrow\ \frac{F_{\bm{\theta}}(x_n)-\hat{\mu}}{\hat{\sigma}}\ (n=1,2,\cdots,10^4)$
	\STATE $\hat{d}(F_{\bm{\theta}}, F^*)\ \leftarrow\ \frac{1}{10^4}\sum^{10^4}_{n=1}(F_{\bm{\theta}}(x_n)-F^*(x_n))^2$
	\FOR{$k=1$ {\bfseries to} $10^4$}
		\IF{$\hat{d}(F_{\bm{\theta}}, F^*)\leq \epsilon_k$}
			\STATE $t_{k}\ \leftarrow\ t_{k}+1$
		\ENDIF
	\ENDFOR
\ENDFOR
\STATE $\hat{R}_{\epsilon_k}(\mathcal{F}, F^*)\ \leftarrow\ \frac{t_{k}}{2\times 10^4}\ (k=1,2,\cdots,10^4)$
\RETURN $\hat{R}_{\epsilon_k}(\mathcal{F}, F^*)\ (k=1,2,\cdots,10^4)$
\end{algorithmic}

\begin{figure}[t]
\begin{center}
\centerline{\includegraphics[height=5cm,width=7cm]{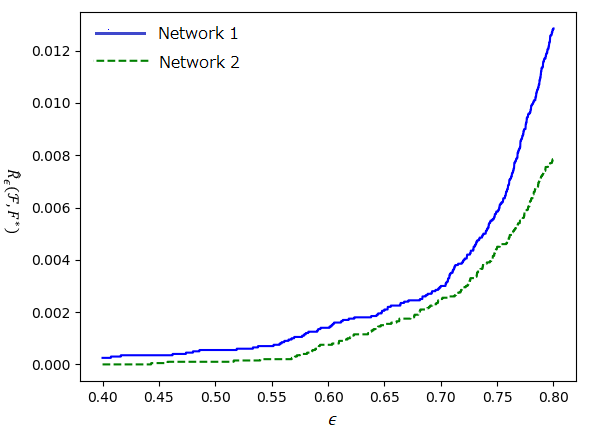}}
\caption{The result when approximately calculating the ratio of the desired parameters with the sin function. From this figure, we can say that to increase layers is more effective than to increase units at each layer on improving on the approximation accuracy since the ratio of the desired parameters of Network 1 is over that of Network 2.}
\label{figresult1}
\end{center}
\end{figure}

\begin{figure}[t]
\begin{center}
\centerline{\includegraphics[height=5.8cm,width=9.5cm]{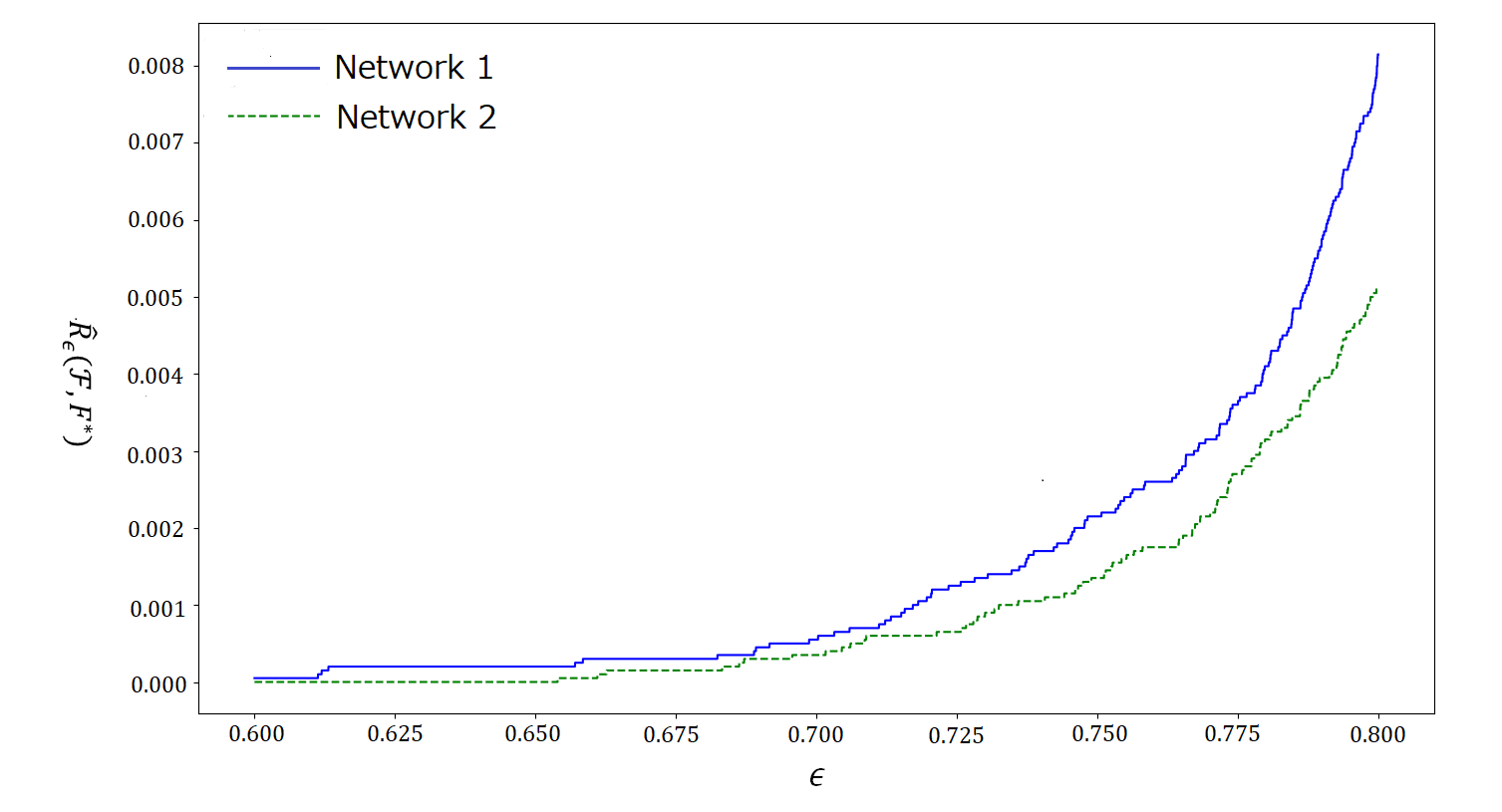}}
\caption{The result when approximately calculating the ratio of the desired parameters with the Weierstrass function. 
From this figure, we can say that to increase layers is more effective than to increase units at each layer on improving on the approximation accuracy since the ratio of the desired parameters of Network 1 is over that of Network 2.
}
\label{figresult2}
\end{center}
\end{figure}

We ran the above algorithm using the neural networks in Table \ref{tabmodel} as $F_{\bm{\theta}}$. Figure \ref{figresult1} and Fig. \ref{figresult2} show the results of the calculation. From these results, we can that Network 1 can express more functions close to $\sin(4\pi x)$ and the Weierstrass function than Network 2. It suggests that to increase layers is more effective than to increase units at each layer on improving the expressivity of neural networks. 
\subsection{The Calculation of the Fineness of Linear Regions}
In order to confirm Theorem \ref{theo}, we approximately calculated the minimum value of the fineness of linear regions $\min_{\bm{\theta}\in\bm{\Theta}}I(\bm{F}_{\bm{\theta}})$ by computer simulations. We must detect linear regions from the input domain of deep neural networks in order to know the fineness of linear regions $I(\bm{F}_{\bm{\theta}})$. Therefore, we did it by calculating the inflection of the derivative of functions computed by deep neural networks. 

We executed the calculation with restricting the input and output dimension to one for simplicity. We generated an arithmetic sequence $x_n=(n-1)\times10^{-5}(n=1,2,\cdots,10^5)$ and calculated the gradient of the line connecting $(x_n, F_{\bm{\theta}}(x_n))$ and $(x_{n+1}, F_{\bm{\theta}}(x_{n+1}))$ such as 
\begin{align}
g_n:=\frac{F_{\bm{\theta}}(x_{n+1})-F_{\bm{\theta}}(x_n)}{x_{n+1}-x_n}. \nonumber
\end{align}
Then we calculated the inflection of $g_{n}$ such as 
\begin{align}
d_{n+1}:=|g_{n+1}-g_{n}|. \nonumber
\end{align}

We regarded that $x_{n+1}$ was the boundary of linear regions if $d_{n+1}$ was larger than a threshold value and calculated the size of the maximal interval between the two boundaries as the fineness of linear regions $I(F_{\bm{\theta}})$. As we have mentioned in Section \ref{secintro}, we must calculate it independently from the efficiency of learning. Therefore, we searched the minimum value of $I(F_{\bm{\theta}})$ by generating the value of $\bm{\theta}$ randomly and repeatedly, not by learning. The following summarizes the algorithm to approximately calculate $\min_{\bm{\theta}\in\bm{\Theta}}I(F_{\bm{\theta}})$.
\begin{algorithmic}
\STATE $x_n\ \leftarrow\ (n-1)\times 10^{-5}\ (n=1,2,\cdots,10^5)$
\STATE $fineness\_list\ \leftarrow\ [\ ]$
\FOR{$i=1$ {\bfseries to} $10^3$}
	\STATE Generate the value of $\bm{\theta}$ by randam sampling of uniform distribution $U(0, 1)$
	\STATE $count\ \leftarrow\ 0$
	\STATE $count\_list\ \leftarrow\ [\ ]$
	\FOR{$n=1$ {\bfseries to} $10^5-2$}
		\FOR{$k=0,1,2$}
			\STATE Compute the output of a neural network $F_{\bm{\theta}}(x_{n+k})$
		\ENDFOR
		\STATE $g_n\ \leftarrow\ \frac{F_{\bm{\theta}}(x_{n+1})-F_{\bm{\theta}}(x_{n})}{x_{n+1}-x_n}$
		\STATE $g_{n+1}\ \leftarrow\ \frac{F_{\bm{\theta}}(x_{n+2})-F_{\bm{\theta}}(x_{n+1})}{x_{n+2}-x_{n+1}}$
		\STATE $d_{n+1}\ \leftarrow\ |g_{n+1}-g_n|$
		\IF{$d_{n+1} < 0.5$}
			\STATE $count\ \leftarrow\ count + 1$
		\ELSE
			\STATE Append $count$ to $count\_list$
			\STATE $count\ \leftarrow\ 0$
		\ENDIF
	\ENDFOR
	\STATE Append $count$ to $count\_list$	
	\STATE $count\_max\ \leftarrow\ $ the maximum value in $count\_list$
	\STATE $fineness\ \leftarrow\ \frac{count\_max}{10^5}$
	\STATE Append $fineness$ to $fineness\_list$
\ENDFOR
\STATE $fineness\_min\ \leftarrow\ $ the minimum value in $fineness\_list$
\RETURN $fineness\_min$
\end{algorithmic}

In the above algorithm, the threshold value of the inflection $d_{n+1}$ is set as 0.5. If $d_{n+1}\geq 0.5$, $x_{n+1}$ is regarded as the boundary of linear regions. The variable $count$ increases if a deep neural network is linear. Then it returns to 0 at the boundary of linear regions. The list $count\_list$ stocks the size of linear regions and the maximum value in $count\_list$ shows the fineness of linear regions. However, the maximum value in $count\_list$ must be divided by $10^5$ because the difference between $x_n$ and $x_{n+1}$ is $10^{-5}$. On the other hand, we also calculated the value of $\prod^{L-1}_{l=1}\lfloor\frac{n_l}{2n_0}\rfloor^{-n_0}$, the right side of (\ref{eqtheo}). 

We ran the above algorithm using the neural networks in Table \ref{tabmodel} as $F_{\bm{\theta}}$. The results of the simulations follow Theorem \ref{theo}. The minimum value of the fineness of linear regions of Network 1 is 0.00004 and that of Network 2 is 0.06345, where the values of $\bm{\theta}$ are restricted to the finite sampling. On the other hand, the value of $\prod^{L-1}_{l=1}\lfloor\frac{n_l}{2n_0}\rfloor^{-n_0}$ of Network 1 is 0.03125 and that of Network 2 is 0.1. From these results, we can confirm Theorem \ref{theo} and say that to increase layers is more effective than to increase units at each layer when we want neural networks to be more flexible.
\section{Conclusion}
In this paper, we proposed two new criteria to evaluate the expressivity of functions computable by deep neural networks independently from the efficiency of learning. The first criterion, the ratio of the desired parameters shows the approximation accuracy of deep neural networks to the target function. The second criterion, the fineness of linear regions shows the property of linear regions of functions computable by deep neural networks. Furthermore, by the two criteria, we showed that to increase layers is more effective than to increase units at each layer on improving the expressivity of deep neural networks. It is hoped that our studies will contribute to a better understanding of the advantage of deepening neural networks.
\section*{Acknowledgment}
This research was supported by JSPS Grants-in-Aid for Scientific Research JP17K00316, JP17K06446, JP18K11585, JP19K04914.
\bibliography{bibdata}
\bibliographystyle{IEEEtran}
\end{document}